\documentclass{article}

\usepackage{microtype}
\usepackage{graphicx}
\usepackage{subfigure}
\usepackage{booktabs} %

\usepackage{amsmath,amsfonts,bm}

\def\eqref#1{equation~\ref{#1}}

\def\1{\bm{1}}

\DeclareMathAlphabet{\mathsfit}{\encodingdefault}{\sfdefault}{m}{sl}
\SetMathAlphabet{\mathsfit}{bold}{\encodingdefault}{\sfdefault}{bx}{n}

\newcommand{\E}{\mathbb{E}}

\usepackage{hyperref}
\usepackage[accepted]{icml2019}
\usepackage{url}
\usepackage{graphicx}
\usepackage{amsmath}
\usepackage{amssymb}
\usepackage{accents}
\usepackage{xcolor}
\DeclareMathOperator*{\mmaximize}{maximize}
\usepackage{pifont}%
\usepackage{wrapfig}
\usepackage{floatflt}

\newcommand{\eat}[1]{}

\newcommand{\lbar}[1]{#1}

\newcommand{\cpc}{\lbar{I}_{\text{NCE}}}
\newcommand{\tcpc}{\lbar{I}_{\text{TNCE}}}
\newcommand{\dv}{\lbar{I}_\text{DV}}
\newcommand{\ba}{\lbar{I}_\text{BA}}
\newcommand{\eba}{\lbar{I}_\text{UBA}}

\newcommand{\inwj}{\lbar{I}_\text{NWJ}}
\newcommand{\nwj}{\lbar{I}_\text{NWJ}}
\newcommand{\ijs}{\lbar{I}_\text{JS}}
\newcommand{\igb}{\lbar{I}_\text{TUBA}}
\newcommand{\interp}{\lbar{I}_\alpha}
\newcommand{\mine}{\lbar{I}_\text{MINE}}
\newcommand{\kl}{\text{KL}}
\newcommand{\xmark}{{\ding{55}}}
\newcommand{\cmark}{{\ding{51}}}

\icmltitlerunning{On Variational Bounds of Mutual Information}

\begin{document}
\twocolumn[
\icmltitle{On Variational Bounds of Mutual Information}

\icmlsetsymbol{equal}{*}

\begin{icmlauthorlist}
\icmlauthor{Ben Poole}{goo}
\icmlauthor{Sherjil Ozair}{goo,mila}
\icmlauthor{A{\"a}ron van den Oord}{dm}
\icmlauthor{Alexander A. Alemi}{goo}
\icmlauthor{George Tucker}{goo}
\end{icmlauthorlist}

\icmlaffiliation{goo}{Google Brain}
\icmlaffiliation{dm}{DeepMind}
\icmlaffiliation{mila}{MILA}
\icmlcorrespondingauthor{Ben Poole}{pooleb@google.com}

\icmlkeywords{Machine Learning, ICML}

\vskip 0.3in
]

\printAffiliationsAndNotice{}  %
\begin{abstract}
Estimating and optimizing Mutual Information~(MI) is core to many problems in machine learning; however, bounding MI in high dimensions is challenging. To establish tractable and scalable objectives, recent work has turned to variational bounds parameterized by neural networks, but the relationships and tradeoffs between these bounds remains unclear. In this work, we unify these recent developments in a single framework. We find that the existing variational lower bounds degrade when the MI is large, exhibiting either high bias or high variance. To address this problem, we introduce a continuum of lower bounds that encompasses previous bounds and flexibly trades off bias and variance. On high-dimensional, controlled problems, we empirically characterize the bias and variance of the bounds and their gradients and demonstrate the effectiveness of our new bounds for estimation and representation learning.
\end{abstract}

\section{Introduction}
Estimating the relationship between pairs of variables is a fundamental problem in science and engineering. Quantifying the degree of the relationship requires a metric that captures a notion of dependency. Here, we focus on mutual information (MI), denoted $I(X; Y)$, which is a reparameterization-invariant measure of dependency:
$$ I(X; Y) = \E_{p(x,y)}\left[ \log \frac{p(x|y)}{p(x)} \right] = \E_{p(x,y)}\left[ \log \frac{p(y|x)}{p(y)} \right]. $$ 
Mutual information estimators are used in computational neuroscience \citep{palmer2015predictive}, Bayesian optimal experimental design \citep{ryan2016review, fostervariational}, understanding neural networks \citep{tishby2000information, tishby2015deep, gabrie2018entropy}, and more.  In practice, estimating MI is challenging as we typically have access to samples but not the underlying distributions~\citep{paninski2003estimation, mcallester2018formal}. Existing sample-based estimators are brittle, with the hyperparameters of the estimator impacting the scientific conclusions~\citep{michael2018on}.%

Beyond estimation, many methods use upper bounds on MI to limit the capacity or contents of representations. For example in the information bottleneck method~\citep{tishby2000information, alemi2016deep}, the representation is optimized to solve a downstream task while being constrained to contain as little information as possible about the input. These techniques have proven useful in a variety of domains, from restricting the capacity of discriminators in GANs~\citep{peng2018variational} to preventing representations from containing information about protected attributes~\citep{moyer2018invariant}.%

Lastly, there are a growing set of methods in representation learning that maximize the mutual information between a learned representation and an aspect of the data. Specifically, given samples from a data distribution, $x \sim p(x)$, the goal is to learn a stochastic representation of the data $p_\theta(y|x)$ that has maximal MI with $X$ subject to constraints on the mapping~\citep[e.g.][]{bell1995information,krause2010discriminative, hu2017learning,oord2018representation,hjelm2018learning,brokenelbo}. To maximize MI, we can compute gradients of a lower bound on MI with respect to the parameters $\theta$ of the stochastic encoder $p_\theta(y|x)$,  which may not require directly estimating MI. %
 \begin{figure}[t!]
    \centering
    \includegraphics[clip, trim={0cm 0.0cm 0cm 0cm},width=0.96\columnwidth]{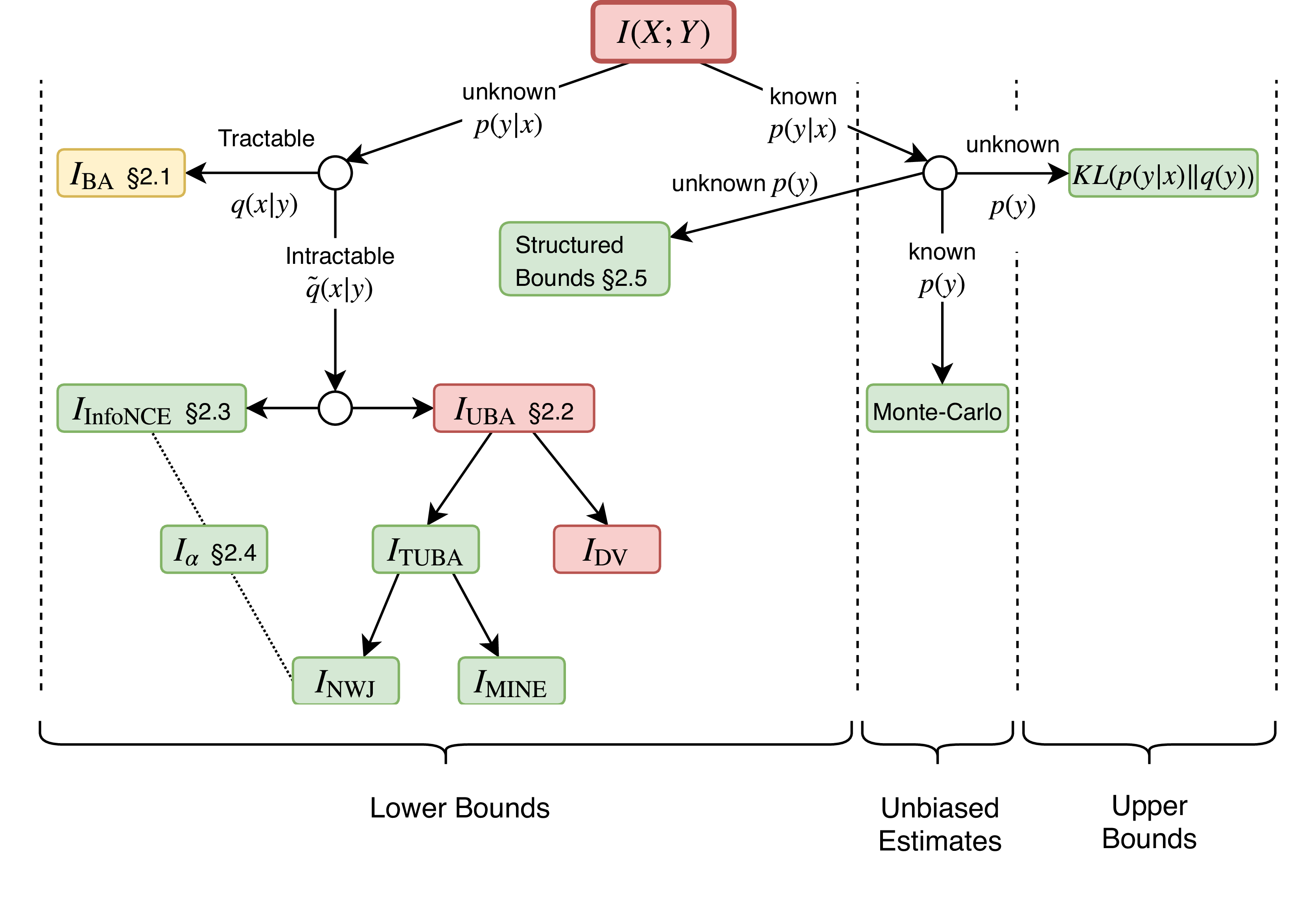}
    \vspace{-9mm}
    \caption{Schematic of variational bounds of mutual information presented in this paper. Nodes are colored based on their tractability for estimation and optimization: green bounds can be used for both, yellow for optimization but not estimation, and red for neither. Children are derived from their parents by introducing new approximations or assumptions. }
    \label{fig:tree}
    \vspace{-.2in}
\end{figure}   
    
\pagebreak %
While many parametric and non-parametric~\citep{nemenman2004entropy,kraskov2004estimating,reshef2011detecting, gao2015efficient} techniques have been proposed to address MI estimation and optimization problems, few of them scale up to the dataset size and dimensionality encountered in modern machine learning problems.

To overcome these scaling difficulties, recent work combines variational bounds~\citep{blei2017variational,donsker1983asymptotic,barber2003algorithm,nguyen2010estimating,fostervariational} with deep learning~\citep{alemi2016deep,brokenelbo,oord2018representation, hjelm2018learning, belghazi2018mutual} to enable differentiable and tractable estimation of mutual information. These papers introduce flexible parametric distributions or \emph{critics} parameterized by neural networks that are used to approximate unkown densities ($p(y)$, $p(y|x)$)  or density ratios ($\frac{p(x|y)}{p(x)}=\frac{p(y|x)}{p(y)}$).   %

In spite of their effectiveness, the properties of existing variational estimators of MI are not well understood.  In this paper, we introduce several results that begin to demystify these approaches and present novel bounds with improved properties (see Fig.~\ref{fig:tree} for a schematic):
\begin{itemize}
        \item We provide a review of existing estimators, discussing their relationships and tradeoffs, including the first proof that the noise contrastive loss in~\citet{oord2018representation} is a lower bound on MI, and that the heuristic ``bias corrected gradients'' in~\citet{belghazi2018mutual} can be justified as unbiased estimates of the gradients of a different lower bound on MI.
        \item We derive a new continuum of multi-sample lower bounds that can flexibly trade off bias and variance, generalizing the bounds of \citep{nguyen2010estimating, oord2018representation}.
        \item We show how to leverage known conditional structure yielding simple lower and upper bounds that sandwich MI in the representation learning context when $p_\theta(y|x)$ is tractable.
        \item We systematically evaluate the bias and variance of MI estimators and their gradients on controlled high-dimensional problems.
        \item We demonstrate the utility of our variational upper and lower bounds in the context of decoder-free disentangled representation learning on dSprites~\citep{dsprites17}.
    \end{itemize}

\section{Variational bounds of MI}%
Here, we review existing variational bounds on MI in a unified framework, and present several new bounds that trade off bias and variance and naturally leverage known conditional densities when they are available. A schematic of the bounds we consider is presented in Fig.~\ref{fig:tree}. 
We begin by reviewing the classic upper and lower bounds of~\citet{barber2003algorithm} and then show how to derive the lower bounds of~\citet{donsker1983asymptotic, nguyen2010estimating, belghazi2018mutual} from an unnormalized variational distribution. Generalizing the unnormalized bounds to the multi-sample setting yields the bound proposed in~\citet{oord2018representation}, and provides the basis for our interpolated bound.

\subsection{Normalized upper and lower bounds}
Upper bounding MI is challenging, but is possible when the conditional distribution $p(y|x)$ is known (e.g. in deep representation learning where $y$ is the stochastic representation). We can build a tractable variational upper bound by introducing a variational approximation $q(y)$ to the intractable marginal $p(y)=\int dx\,p(x)p(y|x)$. By multiplying and dividing the integrand in MI by $q(y)$ and dropping a negative KL term, we get a tractable variational upper bound
~\citep{barber2003algorithm}:
\begin{align}
I(X; Y) &\equiv \E_{p(x,y)} \left[ \log \frac{p(y|x)}{p(y)} \right] \nonumber \\
&= \E_{p(x,y)} \left[ \log \frac{p(y|x)q(y)}{q(y)p(y)} \right] \nonumber \\
&= \E_{p(x,y)}\left[\log \frac{p(y|x)}{q(y)}\right] - KL(p(y)\|q(y))\nonumber\\
&\le \E_{p(x)}\left[KL(p(y|x) \| q(y))\right] \triangleq R,
\label{eq:ba_upper}
\end{align}
which is often referred to as the \emph{rate} in generative models \citep{brokenelbo}. This bound is tight when $q(y)=p(y)$, and requires that computing $\log q(y)$ is tractable. This variational upper bound is often used as a regularizer to limit the capacity of a stochastic representation ~\citep[e.g.][]{rezende2014stochastic, kingma2013auto, burgess2018understanding}. In~\citet{alemi2016deep}, this upper bound is used to prevent the representation from carrying information about the input that is irrelevant for the downstream classification task.

Unlike the upper bound, most variational lower bounds on mutual information do not require direct knowledge of any conditional densities. To establish an initial lower bound on mutual information, we factor MI the opposite direction as the upper bound, and replace the intractable conditional distribution $p(x|y)$ with a tractable optimization problem over a variational distribution $q(x|y)$. As shown in~\citet{barber2003algorithm}, this yields a lower bound on MI due to the non-negativity of the KL divergence:
\begin{equation}
\begin{split}
 I(X; Y) %
 &= \E_{p(x,y)}\left[\log \frac{q(x|y)}{p(x)} \right] \\&\qquad+ \E_{p(y)}\left[ KL(p(x|y)||q(x|y)) \right] \\
 &\ge \E_{p(x,y)}\left[\log q(x|y)\right]  + h(X) \triangleq \ba,
\label{eq:ba}
\end{split}
\end{equation}
where $h(X)$ is the differential entropy of $X$. The bound is tight when $q(x|y) = p(x|y)$, in which case the first term equals the conditional entropy $h(X|Y)$.

Unfortunately, evaluating this objective is generally intractable as the differential entropy of $X$ is often unknown. %
If $h(X)$ is known, this provides a tractable estimate of a lower bound on MI. Otherwise, one can still compare the amount of information different variables (e.g., $Y_1$ and $Y_2$) carry about $X$. 

In the representation learning context where $X$ is data and $Y$ is a learned stochastic representation, the first term of $\ba$ can be thought of as negative reconstruction error or distortion, and the gradient of $\ba$ with respect to the ``encoder'' $p(y|x)$ and variational ``decoder'' $q(x|y)$ is tractable. Thus we can use this objective to learn an encoder $p(y|x)$ that maximizes $I(X;Y)$ as in~\citet{brokenelbo}. However, this approach to representation learning requires building a tractable decoder $q(x|y)$, which is challenging when $X$ is high-dimensional and $h(X|Y)$ is large, for example in video representation learning~\citep{van2016conditional}. 

\subsection{Unnormalized lower bounds}
To derive tractable lower bounds that do not require a tractable decoder, we turn to {\em unnormalized} distributions for the variational family of $q(x|y)$, %
and show how this recovers the estimators of~\citet{donsker1983asymptotic, nguyen2010estimating}.

We choose an energy-based variational family that uses a \emph{critic} $f(x,y)$ and is scaled by the data density $p(x)$:
\begin{equation}
 q(x|y) = \frac{p(x)}{Z(y)} e^{f(x,y)}, \text{ where }
Z(y) = %
\mathbb{E}_{p(x)} \left[ e^{f(x,y)} \right].
    \label{eq:variational_form}
\end{equation}
Substituting this distribution into $\ba$ (Eq.~\ref{eq:ba}) gives a lower bound on MI which we refer to as $\eba$ for the Unnormalized version of the Barber and Agakov bound:
\begin{equation}
 \mathbb{E}_{p(x,y)} \left[f(x,y) \right] - \mathbb{E}_{p(y)} \left[ \log Z(y) \right]  \triangleq \eba.
 \label{eq:unnorm}
\end{equation}
This bound is tight when $f(x,y) = \log p(y|x) + c(y)$, where $c(y)$ is solely a function of $y$ (and not $x$).
Note that by scaling $q(x|y)$ by $p(x)$, the intractable differential entropy term in $\ba$ cancels, but we are still left with an intractable log partition function,  $\log Z(y)$, that prevents evaluation or gradient computation. If we apply Jensen's inequality to $\E_p(y)\left[ \log Z(y) \right]$, we can lower bound Eq.~\ref{eq:unnorm} to recover the bound of~\citet{donsker1983asymptotic}:
\begin{equation}
    \eba \ge 
    \mathbb{E}_{p(x,y)} \left[ f(x,y) \right] - \log \mathbb{E}_{p(y)} \left[  Z(y) \right]  
    \triangleq\dv.
\end{equation}
However, this objective is still intractable. 
Applying Jensen's the other direction by replacing $\log Z(y)=\log \E_{p(x)}\left[e^{f(x,y)}\right]$ with $\E_{p(x)}\left[f(x,y)\right]$ results in a tractable objective, but produces an upper bound on Eq.~\ref{eq:unnorm} (which is itself a lower bound on mutual information). Thus evaluating $\dv$ using a Monte-Carlo approximation of the expectations as in MINE~\citep{belghazi2018mutual} produces \emph{estimates} that are {\em neither an upper or lower bound on MI}. Recent work has studied the convergence and asymptotic consistency of such nested Monte-Carlo estimators, but does not address the problem of building bounds that hold with finite samples \citep{rainforth2018nesting,mathieu18}.

To form a tractable bound,  we can upper bound the log partition function using the inequality: 
$ \log(x) \leq \frac{x}{a} + \log(a) - 1 $
for all $x, a > 0$. Applying this inequality to the second term of Eq.~\ref{eq:unnorm} gives:
$ \log Z(y) \leq \frac{Z(y)}{a(y)} + \log(a(y)) - 1$, 
which is tight when $a(y) = Z(y)$.
This results in a Tractable Unnormalized version of the Barber and Agakov (TUBA) lower bound on MI that admits unbiased estimates and gradients:
\begin{align}
    I\ge \eba
    \geq &~\E_{p(x, y)} \left[f(x,y) \right] \nonumber\\
    &-
\E_{p(y)} \left[ \frac{\E_{p(x)} \left[ e^{f(x , y)} \right]}{a(y)} + \log(a(y)) - 1  \right] \nonumber \\ 
&\triangleq \igb.
    \label{eq:bound_2}
\end{align}
To tighten this lower bound, we maximize with respect to the variational parameters $a(y)$ and $f$. In the InfoMax setting, we can maximize the bound with respect to the stochastic encoder $p_\theta(y|x)$ to increase $I(X; Y)$. Unlike the min-max objective of GANs, all parameters are optimized towards the same objective.

This bound holds for any choice of $a(y) > 0$, with simplifications recovering existing bounds. Letting $a(y)$ be the constant $e$ recovers the bound of Nguyen, Wainwright, and Jordan \citep{nguyen2010estimating} also known as $f$-GAN KL~\citep{fgan} and MINE-$f$~\citep{belghazi2018mutual}\footnote{$\igb$ can also be derived the opposite direction by plugging the critic $f'(x,y) = f(x,y) - \log a(y) + 1$ into $\inwj$.}:
\begin{equation}
\E_{p(x, y)} \left[ f(x,y) \right] - e^{-1}\E_{p(y)} \left[Z(y) \right] \triangleq \inwj.
\end{equation}
This tractable bound no longer requires learning $a(y)$, but now $f(x,y)$ must learn to self-normalize, %
yielding a unique optimal critic $f^*(x,y) = 1+\log\frac{p(x|y)}{p(x)}$.
This requirement of self-normalization is a common choice when learning log-linear models and empirically has been shown not to negatively impact performance \citep{mnih2012fast}.%

Finally, we can set $a(y)$ to be the scalar exponential moving average (EMA) of $e^{f(x,y)}$ across minibatches. This pushes the normalization constant to be independent of $y$, but it no longer has to exactly self-normalize. With this choice of $a(y)$, the gradients of $\igb$ exactly yield the ``improved MINE gradient estimator'' from~\citep{belghazi2018mutual}. This provides sound justification for the heuristic optimization procedure proposed by~\citet{belghazi2018mutual}. %
However, instead of using the critic in the $\dv$ bound to get an estimate that is not a bound on MI as in \citet{belghazi2018mutual}, one can compute an estimate with $\igb$ which results in a valid lower bound. %

To summarize, these unnormalized bounds are attractive because they provide tractable estimators which become tight with the optimal critic. However, in practice they exhibit high variance due to their reliance on high variance upper bounds on the log partition function.

\subsection{Multi-sample unnormalized lower bounds}
To reduce variance, we extend the unnormalized bounds to depend on multiple samples, and show how to recover the low-variance but high-bias MI estimator proposed by~\citet{oord2018representation}.

Our goal is to estimate $I(X_1, Y)$ given samples from $p(x_1)p(y|x_1)$ and access to $K-1$ additional samples $x_{2:K} \sim r^{K-1}(x_{2:K})$ (potentially from a different distribution than $X_1$). For any random variable $Z$ independent from $X$ and $Y$, $I(X,Z; Y) = I(X; Y)$, therefore:
\[
I(X_1;Y) = \E_{r^{K-1}(x_{2:K})}\left[I(X_1; Y)\right] = I\left(X_1, X_{2:K}; Y\right)
\]
This multi-sample mutual information can be estimated using any of the previous bounds, and has the same optimal critic as for $I(X_1;Y)$. For $\nwj$, we have that the optimal critic is $f^*(x_{1:K}, y) = 1 + \log \frac{p(y|x_{1:K})}{p(y)} = 1 + \log \frac{p(y|x_1)}{p(y)} $. However, the critic can now also depend on the additional samples $x_{2:K}$. In particular, setting the critic to $1 + \log \frac{e^{f(x_1, y)}}{a(y; x_{1:K})}$ and $r^{K-1}(x_{2:K}) = \prod_{j=2}^K p(x_j)$, $\nwj$ becomes:
\begin{align}
 I(X_1; Y) \geq 1&+\E_{p(x_{1:K})p(y|x_1)}\left[\log \frac{e^{f(x_1,y)}}{a(y; x_{1:K})}\right] \nonumber\\
&- \E_{p(x_{1:K})p(y)}\left[ \frac{e^{f(x_1,y)}}{a(y; x_{1:K})}\right],
\label{eq:one_multi}   
\end{align}
where we have written the critic using parameters $a(y; x_{1:K})$ to highlight the close connection to the variational parameters in $\igb$. One way to leverage these additional samples from $p(x)$ is to build a Monte-Carlo estimate of the partition function $Z(y)$: 
\[ a(y; x_{1:K}) = m(y; x_{1:K}) = \frac{1}{K}\sum_{i=1}^K e^{f(x_i, y)}.\]
Intriguingly, with this choice, the high-variance term in $\inwj$ that  
estimates an upper bound on $\log Z(y)$ is now upper bounded by $\log K$ as $e^{f(x_1, y)}$ appears in the numerator and also in the denominator (scaled by $\frac{1}{K}$).
If we average the bound over $K$ replicates, reindexing $x_1$ as $x_i$ for each term, then the last term in Eq.~\ref{eq:one_multi} becomes the constant 1:
\begin{align}
&\E_{p(x_{1:K})p(y)}\left[ \frac{e^{f(x_1,y)}}{m(y; x_{1:K})}\right] = \frac{1}{K}\sum_{i=1}^K \E \left[ \frac{e^{f(x_i,y)}}{m(y; x_{1:K})}\right] \nonumber \\
&= \E_{p(x_{1:K})p(y)}\left[  \frac{\frac{1}{K}\sum_{i=1}^K e^{f(x_i,y)}}{m(y; x_{1:K})}\right]=1,
\end{align}
and we exactly recover the lower bound on MI proposed by~\citet{oord2018representation}:
\begin{align}
I(X; Y) \ge \E\left[\frac{1}{K}\sum_{i=1}^K \log \frac{e^{f(x_i, y_i)}}{\frac{1}{K}\sum_{j=1}^K e^{f(x_i,y_j)}}\right]\triangleq \cpc,
\label{eq:cpc}
\end{align}
where the expectation is over $K$ independent samples from the joint distribution: $\prod_j p(x_j, y_j)$. This provides a proof\footnote{The derivation by~\citet{oord2018representation} relied on an approximation, which we show is unnecessary.} that $\cpc$ is a lower bound on MI.  Unlike $\inwj$ where the optimal critic depends on both the conditional and marginal densities, the optimal critic
for $\cpc$ is $f(x,y) = \log p(y|x) + c(y)$ where $c(y)$ is any function that depends on $y$ but not $x$~\citep{ma2018noise}. Thus the critic only has to learn the conditional density and not the marginal density $p(y)$.

As pointed out in~\citet{oord2018representation}, $\cpc$ is upper bounded by $\log K$, meaning that this bound will be loose when $I(X; Y) > \log K$. Although the optimal critic does not depend on the batch size and can be fit with a smaller mini-batches%
, accurately estimating mutual information still needs a large batch size at test time if the mutual information is high.

\subsection{Nonlinearly interpolated lower bounds}
The multi-sample perspective on $\inwj$ allows us to make other choices for the functional form of the critic. Here we propose one simple form for a critic that allows us to nonlinearly interpolate between $\inwj$ and $\cpc$, effectively bridging the gap between the low-bias, high-variance $\inwj$ estimator and the high-bias, low-variance $\cpc$ estimator. Similarly to Eq.~\ref{eq:one_multi}, we set the critic to $1 + \log \frac{e^{f(x_1, y)}}{\alpha m(y; x_{1:K}) + (1 - \alpha)q(y)}$ with $\alpha \in [0, 1]$ to get a continuum of lower bounds:
\begin{align}
1&+\E_{p(x_{1:K})p(y|x_1)}\left[\log \frac{e^{f(x_1,y)}}{\alpha m(y; x_{1:K}) + (1-\alpha)q(y)}\right] \nonumber\\
&- \E_{p(x_{1:K})p(y)}\left[ \frac{e^{f(x_1,y)}}{\alpha m(y; x_{1:K}) + (1-\alpha)q(y)}\right] \triangleq \interp.
\label{eq:interp}
\end{align}
By interpolating between $q(y)$ and $m(y; x_{1:K})$, we can recover $\inwj$ ($\alpha = 0$) or $\cpc$ ($\alpha=1$). Unlike $\cpc$ which is upper bounded by $\log K$, the interpolated bound is upper bounded by $ \log \frac{K}{\alpha}$, allowing us to use $\alpha$ to tune the tradeoff between bias and variance. We can maximize this lower bound in terms of $q(y)$ and $f$. Note that unlike $\cpc$, for $\alpha > 0$ the last term does not vanish and we must sample $y \sim p(y)$ independently from $x_{1:K}$ to form a Monte Carlo approximation for that term. In practice we use a leave-one-out estimate, holding out an element from the minibatch for the independent $y \sim p(y)$ in the second term. We conjecture that the optimal critic for the interpolated bound is achieved when $f(x,y)=\log p(y|x)$ and $q(y)=p(y)$ and use this choice when evaluating the accuracy of the estimates and gradients of $I_\alpha$ with optimal critics.

\subsection{Structured bounds with tractable encoders}
\label{sec:struct}
In the previous sections we presented one variational upper bound and several variational lower bounds. While these bounds are flexible and can make use of any architecture or parameterization for the variational families, we can additionally take into account known problem structure. Here we present several special cases of the previous bounds that can be leveraged when the conditional distribution $p(y|x)$ is known. This case is common in representation learning where $x$ is data and $y$ is a learned stochastic representation.

{\bf InfoNCE with a tractable conditional}.\\
An optimal critic for $\cpc$ is given by $f(x,y)=\log p(y|x)$, so we can simply use the $p(y|x)$ when it is known. This gives us a lower bound on MI without additional variational parameters:
\begin{equation}
I(X; Y) \ge \E \left[ \frac{1}{K}\sum_{i=1}^K \log \frac{p(y_i|x_i)}{\frac{1}{K}\sum_{j=1}^K p(y_i|x_j)}\right],
\label{eq:tcpc}
\end{equation}
where the expectation is over $\prod_j p(x_j, y_j)$. 

{\bf Leave one out upper bound.}\\
Recall that the variational upper bound (Eq.~\ref{eq:ba_upper}) is minimized when our variational $q(y)$ matches the true marginal distribution $p(y) = \int dx\, p(x) p(y|x)$. Given a minibatch of $K$ $(x_i, y_i)$ pairs, we can approximate $p(y) \approx \frac{1}{K} \sum_i p(y | x_i)$~\citep{chen2018isolating}. %
For each example $x_i$ in the minibatch, we can approximate $p(y)$ with the mixture over all other elements: $q_i(y) = \frac{1}{K-1} \sum_{j\ne i} p(y|x_j)$. With this choice of variational distribution, the variational upper bound is:
\begin{equation}
    I(X; Y) \le \E\left[\frac{1}{K}\sum_{i=1}^K \left[\log \frac{p(y_i|x_i)}{\frac{1}{K-1}\sum_{j\ne i}p(y_i|x_j)}\right]\right]
    \label{eq:loo_upper}
\end{equation}
where the expectation is over $\prod_i p(x_i, y_i)$. Combining Eq.~\ref{eq:tcpc} and Eq.~\ref{eq:loo_upper}, we can sandwich MI without introducing learned variational distributions. Note that the only difference between these bounds is whether $p(y_i|x_i)$ is included in the denominator. 
Similar mixture distributions have been used in prior work but they require additional parameters~\citep{pmlr-v84-tomczak18a,kolchinsky2017nonlinear}. 

{\bf Reparameterizing critics.}\\
For $\inwj$, the optimal critic is given by $1+\log \frac{p(y|x)}{p(y)}$, so it is possible to use a critic $f(x,y) = 1 + \log \frac{p(y|x)}{q(y)}$ and optimize only over $q(y)$ when $p(y|x)$ is known. The resulting bound resembles the variational upper bound (Eq.~\ref{eq:ba_upper}) with a correction term to make it a lower bound:
\begin{align}
I &\ge \E_{p(x,y)}\left[\log \frac{p(y|x)}{q(y)}\right] - \E_{p(y)}\left[\frac{\E_{p(x)}\left[p(y|x)\right]}{q(y)}\right] + 1 \nonumber \\
&= R +1- \E_{p(y)}\left[\frac{\E_{p(x)}\left[p(y|x)\right]}{q(y)}\right]
\label{eq:tnwj}
\end{align}
This bound is valid for any choice of $q(y)$, including unnormalized $q$.

Similarly, for the interpolated bounds we can use $f(x,y)=\log p(y|x)$ and only optimize over the $q(y)$ in the denominator. In practice, we find reparameterizing the critic to be beneficial as the critic no longer needs to learn the mapping between $x$ and $y$, and instead only has to learn an approximate marginal $q(y)$ in the typically lower-dimensional representation space.
\begin{figure*}[t]
    \centering
    \includegraphics[width=\textwidth]{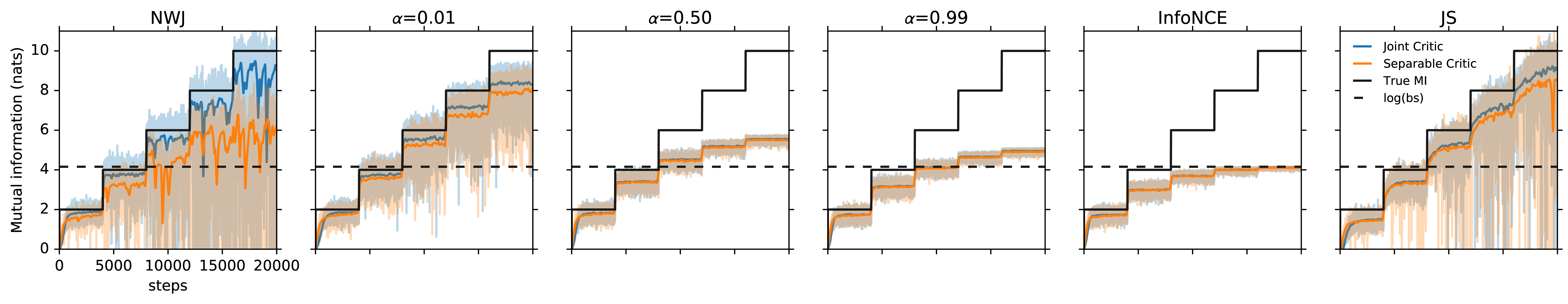}
    \includegraphics[clip, trim=0cm 0cm 0cm 0.7cm,width=\textwidth]{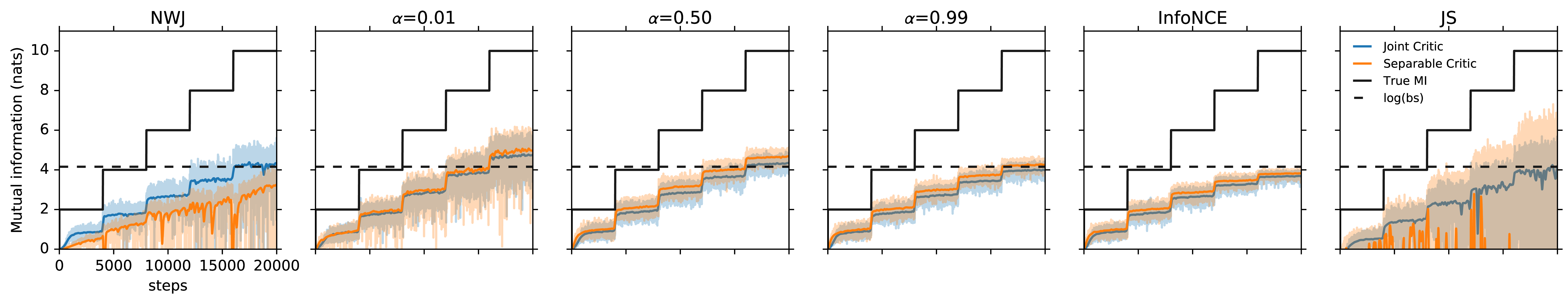}
    \vspace{-.3in}
    \caption{Performance of bounds at estimating mutual information. {\bf Top}: The dataset $p(x,y; \rho)$ is a correlated Gaussian with the correlation $\rho$ stepping over time. {\bf Bottom}: the dataset is created by drawing $x,y\sim p(x, y; \rho)$ and then transforming $y$ to get $(Wy)^3$ where $W_{ij} \sim \mathcal{N}(0, 1)$ and the cubing is elementwise. Critics are trained to maximize each lower bound on MI, and the objective (light) and smoothed objective (dark) are plotted for each technique and critic type. The single-sample bounds ($\nwj$ and $\ijs$) have higher variance than $\cpc$ and $\interp$, but achieve competitive estimates on both datasets. While $\cpc$ is a poor estimator of MI with the small training batch size of 64, the interpolated bounds are able to provide less biased estimates than $\cpc$ with less variance than $\nwj$. For the more challenging nonlinear relationship in the bottom set of panels, the best estimates of MI are with $\alpha=0.01$. Using a joint critic (orange) outperforms a separable critic (blue) for $\inwj$ and $\ijs$, while the multi-sample bounds are more robust to the choice of critic architecture. }
    \label{fig:estimation}
\end{figure*}

{\bf Upper bounding total correlation.}\\
Minimizing statistical dependency in representations is a common goal in disentangled representation learning. Prior work has focused on two approaches that both minimize lower bounds: (1) using adversarial learning~\citep{kim2018disentangling, hjelm2018learning}, or (2) using minibatch approximations where again a lower bound is minimized~\citep{chen2018isolating}.  To measure and minimize statistical dependency, we would like an upper bound, not a lower bound. In the case of a mean field encoder $p(y|x)=\prod_ip(y_i|x)$, we can factor the total correlation into two information terms, and form a tractable upper bound. First, we can write the total correlation as: $TC(Y) = \sum_i I(X; Y_i) - I(X; Y)$. We can then use either the standard (Eq.~\ref{eq:ba_upper}) or the leave one out upper bound (Eq.~\ref{eq:loo_upper}) for each term in the summation, and any of the lower bounds for $I(X; Y)$. If $I(X; Y)$ is small, we can use the leave one out upper bound (Eq.~\ref{eq:loo_upper}) and $\cpc$ (Eq.~\ref{eq:tcpc}) for the lower bound and get a tractable upper bound on total correlation without any variational distributions or critics. Broadly, we can convert lower bounds on mutual information into upper bounds on KL divergences when the conditional distribution is tractable.

\subsection{From density ratio estimators to bounds}

Note that the optimal critic for both $\inwj$ and $\cpc$ are functions of the log density ratio $\log \frac{p(y|x)}{p(y)}$. So, given a log density ratio estimator, we can estimate the optimal critic and form a lower bound on MI.  In practice, we find that training a critic using the Jensen-Shannon divergence (as in~\citet{fgan, hjelm2018learning}), yields an estimate of the log density ratio that is lower variance and as accurate as training with $\nwj$. Empirically we find that training the critic using gradients of $\inwj$ can be unstable due to the $\exp$ from the upper bound on the log partition function in the $\inwj$ objective. Instead, one can train a log density ratio estimator to maximize a lower bound on the Jensen-Shannon (JS) divergence, and use the density ratio estimate in $\inwj$ (see Appendix~\ref{app:ijs} for details). We call this approach $\ijs$ as we update the critic using the JS as in~\citep{hjelm2018learning}, but still compute a MI lower bound with $\inwj$. %
This approach is similar to~\citep{poole2016, mescheder2017adversarial} but results in a bound instead of an unbounded estimate based on a Monte-Carlo approximation of the $f$-divergence.%

\section{Experiments}
First, we evaluate the performance of MI bounds on two simple tractable toy problems. 
Then, we conduct a more thorough analysis of the bias/variance tradeoffs in MI estimates and gradient estimates given the optimal critic.
Our goal in these experiments was to verify the theoretical results
in Section 2, and show that the interpolated bounds can achieve better estimates of MI when the relationship between the variables is nonlinear. Finally, we highlight the utility of these bounds for disentangled representation learning on the dSprites datasets.

{\bf Comparing estimates across different lower bounds.}\\
We applied our estimators to two different toy problems, (1) a correlated Gaussian problem taken from~\citet{belghazi2018mutual} where $(x, y)$ are drawn from a 20-d Gaussian distribution with correlation $\rho$ (see Appendix~\ref{app:toy} for details), and we vary $\rho$ over time, and (2) the same as in (1) but we apply a random linear transformation followed by a cubic nonlinearity to $y$ to get samples $(x, (Wy)^3)$. As long as the linear transformation is full rank,  $I(X; Y) = I(X; (WY)^3)$. We find that the single-sample unnormalized critic estimates of MI exhibit high variance, and are challenging to tune for even these problems. In congtrast, the multi-sample estimates of $\cpc$ are low variance, but have estimates that saturate at $\log(\text{batch size})$. The interpolated bounds trade off bias for variance, and achieve the best estimates of MI for the second problem. None of the estimators exhibit low variance {\em and} good estimates of MI at high rates, supporting the theoretical findings of~\citet{mcallester2018formal}.

{\bf Efficiency-accuracy tradeoffs for critic architectures.} One major difference between the critic architectures used in~\citep{oord2018representation} and~\citep{belghazi2018mutual} is the structure of the critic architecture.~\citet{oord2018representation} uses a separable critic $f(x,y)=h(x)^Tg(y)$ which requires only $2N$ forward passes through a neural network for a batch size of $N$. However,~\citet{belghazi2018mutual} use a joint critic, where $x,y$ are concatenated and fed as input to one network, thus requiring $N^2$ forward passes. For both toy problems, we found that separable critics (orange) increased the variance of the estimator and generally performed worse than joint critics (blue) when using $\inwj$ or $\ijs$ (Fig.~\ref{fig:estimation}). However, joint critics scale poorly with batch size, and it is possible that separable critics require larger neural networks to get similar performance. %

\begin{figure}[t!]
    \centering
    \includegraphics[width=1\columnwidth]{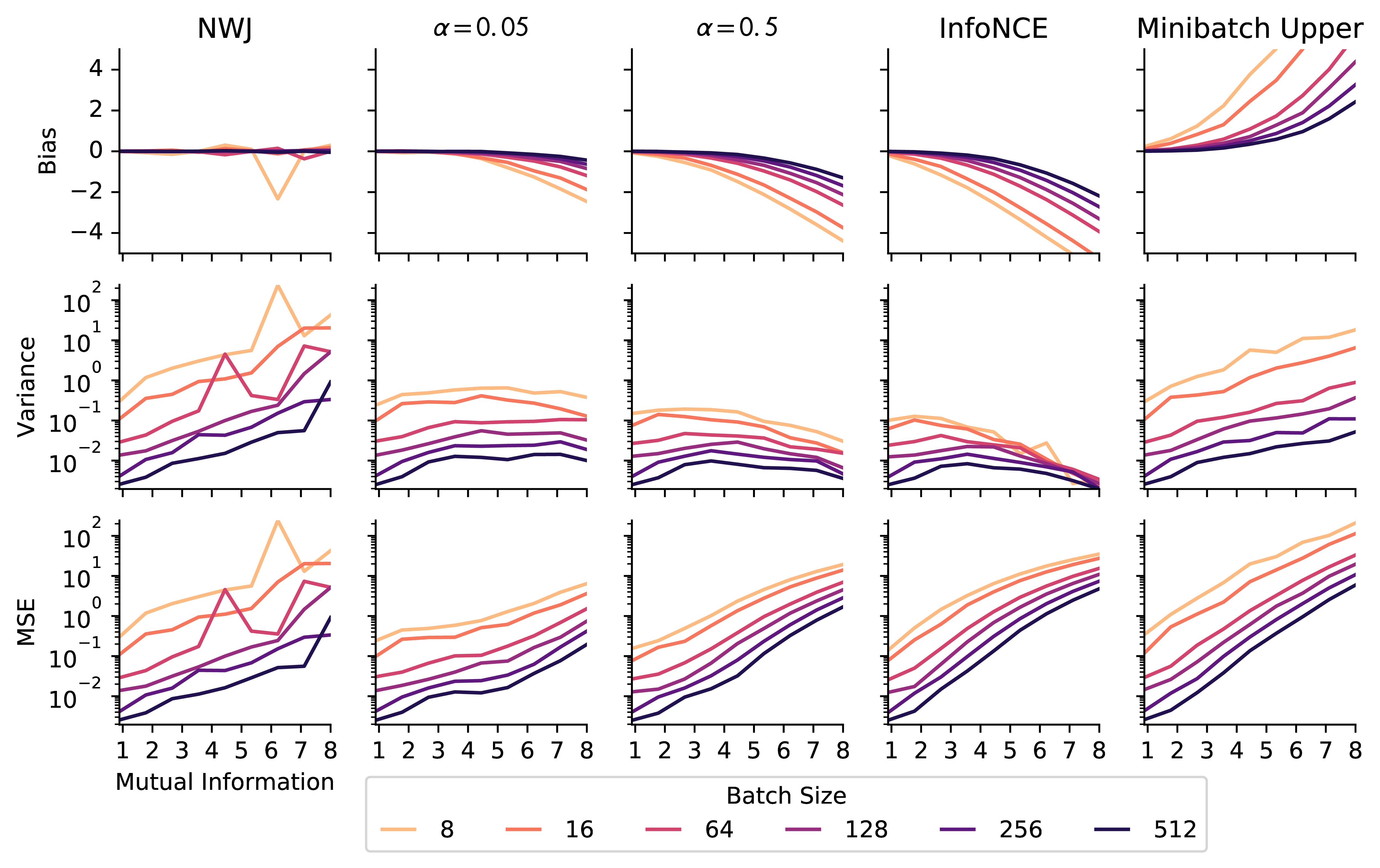}
    \caption{{\bf Bias and variance of MI estimates with the optimal critic.} While $\nwj$ is unbiased when given the optimal critic, $\cpc$ can exhibit large bias that grows linearly with MI. The $\interp$ bounds trade off bias and variance to recover more accurate bounds in terms of MSE in certain regimes. }
    \label{fig:estbiasvar}
\end{figure}
{\bf Bias-variance tradeoff for optimal critics.}\\
To better understand the behavior of different estimators, we analyzed the bias and variance of each estimator as a function of batch size given the {\em optimal critic} (Fig.~\ref{fig:estbiasvar}). We again evaluated the estimators on the 20-d correlated Gaussian distribution and varied $\rho$ to achieve different values of MI. While $\nwj$ is an unbiased estimator of MI, it exhibits high variance when the MI is large and the batch size is small. As noted in~\citet{oord2018representation}, the $\cpc$ estimate is upper bounded by $\log(\text{batch size})$. This results in high bias but low variance when the batch size is small and the MI is large. In this regime, the absolute value of the bias grows linearly with MI because the objective saturates to a constant while the MI continues to grow linearly. In contrast, the $\interp$ bounds are less biased than $\cpc$ and lower variance than $\nwj$, resulting in a mean squared error (MSE) that can be smaller than either $\nwj$ or $\cpc$. We can also see that the leave one out upper bound (Eq.~\ref{eq:loo_upper}) has large bias and variance when the batch size is too small.

\begin{figure}[t!]
    \centering
    \includegraphics[width=1.0\columnwidth]{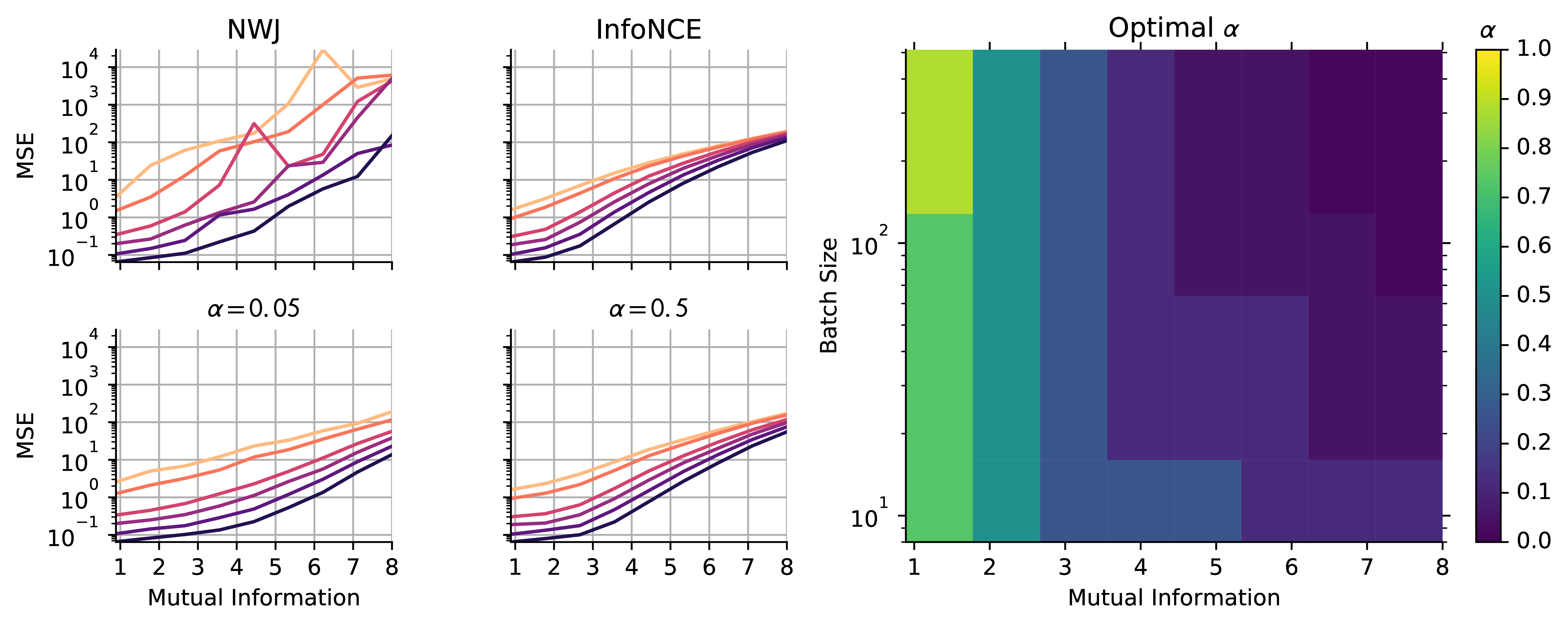}
    \caption{{\bf Gradient accuracy of MI estimators.} {\bf Left:} MSE between the true encoder gradients and approximate gradients as a function of mutual information and batch size (colors the same as in Fig.~\ref{fig:estbiasvar} ). {\bf Right:} For each mutual information and batch size, we evaluated the $I_\alpha$ bound with different $\alpha$s and found the $\alpha$ that had the smallest gradient MSE. For small MI and small size, $\cpc$-like objectives are preferred, while for large MI and large batch size, $\nwj$-like objectives are preferred.
    }
    \label{fig:gradbiasvar}
\end{figure}

{\bf Bias-variance tradeoffs for representation learning.}\\
To better understand whether the bias and variance of the estimated MI impact representation learning, we 
looked at the accuracy of the gradients of the estimates with respect to a stochastic encoder $p(y|x)$ versus the true gradient of MI with respect to the encoder. In order to have access to ground truth gradients, we restrict our model to $p_\rho(y_i|x_i) =\mathcal{N}(\rho_i x, \sqrt{1-\rho_i^2})$ where we have a separate correlation parameter for each dimension $i$, and look at the gradient of MI with respect to the vector of parameters $\rho$. We evaluate the accuracy of the gradients by computing the MSE between the true and approximate gradients. 
For different settings of the parameters $\rho$, we identify which $\alpha$ performs best as a function of batch size and mutual information. In Fig.~\ref{fig:gradbiasvar}, we show that the optimal $\alpha$ for the interpolated bounds depends strongly on batch size and the true mutual information. For smaller batch sizes and MIs, $\alpha$ close to 1 ($\cpc)$ is preferred, while for larger batch sizes and MIs, $\alpha$ closer to 0 ($\nwj$) is preferred. The reduced gradient MSE of the $\interp$ bounds points to their utility as an objective for training encoders in the InfoMax setting.

\subsection{Decoder-free representation learning on dSprites}
Many recent papers in representation learning have focused on learning latent representations in a generative model that correspond to human-interpretable or ``disentangled'' concepts~\citep{higgins2016early, burgess2018understanding,chen2018isolating, kumar2017variational}. While the exact definition of disentangling remains elusive~\citep{locatello2018challenging, higgins18, mathieu18}, many papers have focused on reducing statistical dependency between latent variables as a proxy~\citep{kim2018disentangling, chen2018isolating, kumar2017variational}. Here we show how a decoder-free information maximization approach subject to smoothness and independence constraints can retain much of the representation learning capabilities of latent-variable generative models on the dSprites dataset (a 2d dataset of white shapes on a black background with varying shape, rotation, scale, and position from \citet{dsprites17}). 

To estimate and maximize the information contained in the representation $Y$ about the input $X$, we use the $\ijs$ lower bound, with a structured critic that leverages the known stochastic encoder $p(y|x)$ but learns an unnormalized variational approximation $q(y)$ to the prior. To encourage independence, we form an upper bound on the total correlation of the representation, $TC(Y)$, by leveraging our novel variational bounds. In particular, we reuse the $\ijs$ lower bound of $I(X; Y)$, and use the leave one out upper bounds (Eq.~\ref{eq:loo_upper}) for each $I(X; Y_i)$. Unlike prior work in this area with VAEs,~\citep{kim2018disentangling, chen2018isolating, hjelm2018learning, kumar2017variational},
this approach tractably estimates and removes statistical dependency in the representation without resorting to adversarial techniques, moment matching, or minibatch lower bounds in the wrong direction. %

As demonstrated in~\citet{krause2010discriminative}, information maximization alone is ineffective at learning useful representations from finite data. Furthermore, minimizing statistical dependency is also insufficient, as we can always find an invertible function that maintains the same amount of information and correlation structure, but scrambles the representation~\citep{locatello2018challenging}. We can avoid these issues by introducing additional inductive biases into the representation learning problem. In particular, here we add a simple smoothness regularizer that forces nearby points in $x$ space to be mapped to similar regions in $y$ space: $R(\theta) = KL(p_\theta(y|x) \| p_\theta(y|x+\epsilon))$ where $\epsilon \sim \mathcal{N}(0, 0.5)$. 

The resulting regularized InfoMax objective we optimize is:
\begin{align}
\mmaximize_{p(y|x)}\; &I(X; Y) \\[-10pt]
\text{subject to }\; &TC(Y) = \sum_{i=1}^K I(X; Y_i) - I(X; Y) \le \delta\nonumber\\
&\mathbb{E}_{p(x)p(\epsilon)}\left[KL(p(y|x) \| p(y|x+\epsilon))\right] \le \gamma\nonumber
\end{align}
We use the convolutional encoder architecture from~\citet{burgess2018understanding, locatello2018challenging} for $p(y|x)$, and a two hidden layer fully-connected neural network to parameterize the unnormalized variational marginal $q(y)$ used by $\ijs$.

\begin{figure}[t!]
    \centering
    \includegraphics[width=0.7\columnwidth]{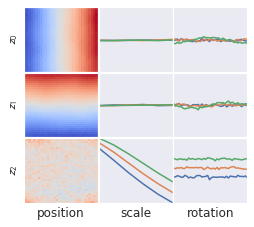}
    \caption{{\bf Feature selectivity on dSprites.} The representation learned with our regularized InfoMax objective exhibits disentangled features for position and scale, but not rotation. Each row corresponds to a different active latent dimension. The first column depicts the position tuning of the latent variable, where the x and y axis correspond to x/y position, and the color corresponds to the average activation of the latent variable in response to an input at that position (red is high, blue is low). The scale and rotation columns show the average value of the latent on the $y$ axis, and the value of the ground truth factor (scale or rotation) on the x axis. %
    }
    \label{fig:dsprites}
\end{figure}

Empirically, we find that this variational regularized infomax objective is able to learn x and y position, and scale, but not rotation (Fig.~\ref{fig:dsprites}, see \citet{chen2018isolating} for more details on the visualization).  To the best of our knowledge, the only other decoder-free representation learning result on dSprites is~\citet{pfau2018minimally}, which recovers shape and rotation but not scale on a simplified version of the dSprites dataset with one shape.

\section{Discussion}
In this work, we reviewed and presented several new bounds on mutual information. We showed that our new interpolated bounds are able to trade off bias for variance to yield better estimates of MI. However, none of the approaches we considered here are capable of providing low-variance, low-bias estimates when the MI is large and the batch size is small. Future work should identify whether such estimators are impossible~\citep{mcallester2018formal}, or whether certain distributional assumptions or neural network inductive biases can be leveraged to build tractable estimators. Alternatively, it may be easier to estimate gradients of MI than estimating MI. For example, maximizing $\ba$ is feasible even though we do not have access to the constant data entropy. There may be better approaches in this setting when we do not care about MI estimation and only care about computing gradients of MI for minimization or maximization.

A limitation of our analysis and experiments is that they focus on the regime where the dataset is infinite and there is no overfitting. In this setting, we do not have to worry about differences in MI on training vs. heldout data, nor do we have to tackle biases of finite samples. Addressing and understanding this regime is an important area for future work.

Another open question is whether mutual information maximization is a more useful objective for representation learning than other unsupervised or self-supervised approaches \citep{noroozi2016unsupervised, doersch2015unsupervised, dosovitskiy2014discriminative}. %
While deviating from mutual information maximization loses a number of connections to information theory, it may provide other mechanisms for learning features that are useful for downstream tasks. In future work, we hope to evaluate these estimators on larger-scale representation learning tasks to address these questions.
\bibliography{icml}

\begin{thebibliography}{51}
\providecommand{\natexlab}[1]{#1}
\providecommand{\url}[1]{\texttt{#1}}
\expandafter\ifx\csname urlstyle\endcsname\relax
  \providecommand{\doi}[1]{doi: #1}\else
  \providecommand{\doi}{doi: \begingroup \urlstyle{rm}\Url}\fi

\bibitem[Alemi et~al.(2016)Alemi, Fischer, Dillon, and Murphy]{alemi2016deep}
Alemi, A.~A., Fischer, I., Dillon, J.~V., and Murphy, K.
\newblock Deep variational information bottleneck.
\newblock \emph{arXiv preprint arXiv:1612.00410}, 2016.

\bibitem[Alemi et~al.(2017)Alemi, Poole, Fischer, Dillon, Saurous, and
  Murphy]{brokenelbo}
Alemi, A.~A., Poole, B., Fischer, I., Dillon, J.~V., Saurous, R.~A., and
  Murphy, K.
\newblock Fixing a broken elbo, 2017.

\bibitem[Barber \& Agakov(2003)Barber and Agakov]{barber2003algorithm}
Barber, D. and Agakov, F.
\newblock The im algorithm: A variational approach to information maximization.
\newblock In \emph{NIPS}, pp.\  201--208. MIT Press, 2003.

\bibitem[Belghazi et~al.(2018)Belghazi, Baratin, Rajeshwar, Ozair, Bengio,
  Hjelm, and Courville]{belghazi2018mutual}
Belghazi, M.~I., Baratin, A., Rajeshwar, S., Ozair, S., Bengio, Y., Hjelm, D.,
  and Courville, A.
\newblock Mutual information neural estimation.
\newblock In \emph{International Conference on Machine Learning}, pp.\
  530--539, 2018.

\bibitem[Bell \& Sejnowski(1995)Bell and Sejnowski]{bell1995information}
Bell, A.~J. and Sejnowski, T.~J.
\newblock An information-maximization approach to blind separation and blind
  deconvolution.
\newblock \emph{Neural computation}, 7\penalty0 (6):\penalty0 1129--1159, 1995.

\bibitem[Blei et~al.(2017)Blei, Kucukelbir, and McAuliffe]{blei2017variational}
Blei, D.~M., Kucukelbir, A., and McAuliffe, J.~D.
\newblock Variational inference: A review for statisticians.
\newblock \emph{Journal of the American Statistical Association}, 112\penalty0
  (518):\penalty0 859--877, 2017.

\bibitem[Burgess et~al.(2018)Burgess, Higgins, Pal, Matthey, Watters,
  Desjardins, and Lerchner]{burgess2018understanding}
Burgess, C.~P., Higgins, I., Pal, A., Matthey, L., Watters, N., Desjardins, G.,
  and Lerchner, A.
\newblock Understanding disentangling in $\beta$-vae.
\newblock \emph{arXiv preprint arXiv:1804.03599}, 2018.

\bibitem[Chen et~al.(2018)Chen, Li, Grosse, and Duvenaud]{chen2018isolating}
Chen, T.~Q., Li, X., Grosse, R., and Duvenaud, D.
\newblock Isolating sources of disentanglement in variational autoencoders.
\newblock \emph{arXiv preprint arXiv:1802.04942}, 2018.

\bibitem[Doersch et~al.(2015)Doersch, Gupta, and
  Efros]{doersch2015unsupervised}
Doersch, C., Gupta, A., and Efros, A.~A.
\newblock Unsupervised visual representation learning by context prediction.
\newblock In \emph{Proceedings of the IEEE International Conference on Computer
  Vision}, pp.\  1422--1430, 2015.

\bibitem[Donsker \& Varadhan(1983)Donsker and Varadhan]{donsker1983asymptotic}
Donsker, M.~D. and Varadhan, S.~S.
\newblock Asymptotic evaluation of certain markov process expectations for
  large time. iv.
\newblock \emph{Communications on Pure and Applied Mathematics}, 36\penalty0
  (2):\penalty0 183--212, 1983.

\bibitem[Dosovitskiy et~al.(2014)Dosovitskiy, Springenberg, Riedmiller, and
  Brox]{dosovitskiy2014discriminative}
Dosovitskiy, A., Springenberg, J.~T., Riedmiller, M., and Brox, T.
\newblock Discriminative unsupervised feature learning with convolutional
  neural networks.
\newblock In \emph{Advances in Neural Information Processing Systems}, pp.\
  766--774, 2014.

\bibitem[Foster et~al.(2018)Foster, Jankowiak, Bingham, Teh, Rainforth, and
  Goodman]{fostervariational}
Foster, A., Jankowiak, M., Bingham, E., Teh, Y.~W., Rainforth, T., and Goodman,
  N.
\newblock Variational optimal experiment design: Efficient automation of
  adaptive experiments.
\newblock In \emph{NeurIPS Bayesian Deep Learning Workshop}, 2018.

\bibitem[Gabri{\'e} et~al.(2018)Gabri{\'e}, Manoel, Luneau, Barbier, Macris,
  Krzakala, and Zdeborov{\'a}]{gabrie2018entropy}
Gabri{\'e}, M., Manoel, A., Luneau, C., Barbier, J., Macris, N., Krzakala, F.,
  and Zdeborov{\'a}, L.
\newblock Entropy and mutual information in models of deep neural networks.
\newblock \emph{arXiv preprint arXiv:1805.09785}, 2018.

\bibitem[Gao et~al.(2015)Gao, Ver~Steeg, and Galstyan]{gao2015efficient}
Gao, S., Ver~Steeg, G., and Galstyan, A.
\newblock Efficient estimation of mutual information for strongly dependent
  variables.
\newblock In \emph{Artificial Intelligence and Statistics}, pp.\  277--286,
  2015.

\bibitem[Higgins et~al.(2016)Higgins, Matthey, Glorot, Pal, Uria, Blundell,
  Mohamed, and Lerchner]{higgins2016early}
Higgins, I., Matthey, L., Glorot, X., Pal, A., Uria, B., Blundell, C., Mohamed,
  S., and Lerchner, A.
\newblock Early visual concept learning with unsupervised deep learning.
\newblock \emph{arXiv preprint arXiv:1606.05579}, 2016.

\bibitem[Higgins et~al.(2018)Higgins, Amos, Pfau, Racani{\`{e}}re, Matthey,
  Rezende, and Lerchner]{higgins18}
Higgins, I., Amos, D., Pfau, D., Racani{\`{e}}re, S., Matthey, L., Rezende,
  D.~J., and Lerchner, A.
\newblock Towards a definition of disentangled representations.
\newblock \emph{CoRR}, abs/1812.02230, 2018.

\bibitem[Hjelm et~al.(2018)Hjelm, Fedorov, Lavoie-Marchildon, Grewal,
  Trischler, and Bengio]{hjelm2018learning}
Hjelm, R.~D., Fedorov, A., Lavoie-Marchildon, S., Grewal, K., Trischler, A.,
  and Bengio, Y.
\newblock Learning deep representations by mutual information estimation and
  maximization.
\newblock \emph{arXiv preprint arXiv:1808.06670}, 2018.

\bibitem[Hu et~al.(2017)Hu, Miyato, Tokui, Matsumoto, and
  Sugiyama]{hu2017learning}
Hu, W., Miyato, T., Tokui, S., Matsumoto, E., and Sugiyama, M.
\newblock Learning discrete representations via information maximizing
  self-augmented training.
\newblock \emph{arXiv preprint arXiv:1702.08720}, 2017.

\bibitem[Kim \& Mnih(2018)Kim and Mnih]{kim2018disentangling}
Kim, H. and Mnih, A.
\newblock Disentangling by factorising.
\newblock \emph{arXiv preprint arXiv:1802.05983}, 2018.

\bibitem[Kingma \& Welling(2013)Kingma and Welling]{kingma2013auto}
Kingma, D.~P. and Welling, M.
\newblock Auto-encoding variational bayes.
\newblock \emph{nternational Conference on Learning Representations}, 2013.

\bibitem[Kolchinsky et~al.(2017)Kolchinsky, Tracey, and
  Wolpert]{kolchinsky2017nonlinear}
Kolchinsky, A., Tracey, B.~D., and Wolpert, D.~H.
\newblock Nonlinear information bottleneck.
\newblock \emph{arXiv preprint arXiv:1705.02436}, 2017.

\bibitem[Kraskov et~al.(2004)Kraskov, St{\"o}gbauer, and
  Grassberger]{kraskov2004estimating}
Kraskov, A., St{\"o}gbauer, H., and Grassberger, P.
\newblock Estimating mutual information.
\newblock \emph{Physical review E}, 69\penalty0 (6):\penalty0 066138, 2004.

\bibitem[Krause et~al.(2010)Krause, Perona, and
  Gomes]{krause2010discriminative}
Krause, A., Perona, P., and Gomes, R.~G.
\newblock Discriminative clustering by regularized information maximization.
\newblock In \emph{Advances in neural information processing systems}, pp.\
  775--783, 2010.

\bibitem[Kumar et~al.(2017)Kumar, Sattigeri, and
  Balakrishnan]{kumar2017variational}
Kumar, A., Sattigeri, P., and Balakrishnan, A.
\newblock Variational inference of disentangled latent concepts from unlabeled
  observations.
\newblock \emph{arXiv preprint arXiv:1711.00848}, 2017.

\bibitem[Locatello et~al.(2018)Locatello, Bauer, Lucic, Gelly, Sch{\"o}lkopf,
  and Bachem]{locatello2018challenging}
Locatello, F., Bauer, S., Lucic, M., Gelly, S., Sch{\"o}lkopf, B., and Bachem,
  O.
\newblock Challenging common assumptions in the unsupervised learning of
  disentangled representations.
\newblock \emph{arXiv preprint arXiv:1811.12359}, 2018.

\bibitem[Ma \& Collins(2018)Ma and Collins]{ma2018noise}
Ma, Z. and Collins, M.
\newblock Noise contrastive estimation and negative sampling for conditional
  models: Consistency and statistical efficiency.
\newblock \emph{arXiv preprint arXiv:1809.01812}, 2018.

\bibitem[Mathieu et~al.(2018)Mathieu, Rainforth, Siddharth, and Teh]{mathieu18}
Mathieu, E., Rainforth, T., Siddharth, N., and Teh, Y.~W.
\newblock Disentangling disentanglement in variational auto-encoders, 2018.

\bibitem[Matthey et~al.(2017)Matthey, Higgins, Hassabis, and
  Lerchner]{dsprites17}
Matthey, L., Higgins, I., Hassabis, D., and Lerchner, A.
\newblock dsprites: Disentanglement testing sprites dataset.
\newblock https://github.com/deepmind/dsprites-dataset/, 2017.

\bibitem[McAllester \& Stratos(2018)McAllester and
  Stratos]{mcallester2018formal}
McAllester, D. and Stratos, K.
\newblock Formal limitations on the measurement of mutual information, 2018.

\bibitem[Mescheder et~al.(2017)Mescheder, Nowozin, and
  Geiger]{mescheder2017adversarial}
Mescheder, L., Nowozin, S., and Geiger, A.
\newblock Adversarial variational bayes: Unifying variational autoencoders and
  generative adversarial networks.
\newblock \emph{arXiv preprint arXiv:1701.04722}, 2017.

\bibitem[Mnih \& Teh(2012)Mnih and Teh]{mnih2012fast}
Mnih, A. and Teh, Y.~W.
\newblock A fast and simple algorithm for training neural probabilistic
  language models.
\newblock \emph{arXiv preprint arXiv:1206.6426}, 2012.

\bibitem[Moyer et~al.(2018)Moyer, Gao, Brekelmans, Galstyan, and
  Ver~Steeg]{moyer2018invariant}
Moyer, D., Gao, S., Brekelmans, R., Galstyan, A., and Ver~Steeg, G.
\newblock Invariant representations without adversarial training.
\newblock In \emph{Advances in Neural Information Processing Systems}, pp.\
  9102--9111, 2018.

\bibitem[Nemenman et~al.(2004)Nemenman, Bialek, and van
  Steveninck]{nemenman2004entropy}
Nemenman, I., Bialek, W., and van Steveninck, R. d.~R.
\newblock Entropy and information in neural spike trains: Progress on the
  sampling problem.
\newblock \emph{Physical Review E}, 69\penalty0 (5):\penalty0 056111, 2004.

\bibitem[Nguyen et~al.(2010)Nguyen, Wainwright, and
  Jordan]{nguyen2010estimating}
Nguyen, X., Wainwright, M.~J., and Jordan, M.~I.
\newblock Estimating divergence functionals and the likelihood ratio by convex
  risk minimization.
\newblock \emph{IEEE Transactions on Information Theory}, 56\penalty0
  (11):\penalty0 5847--5861, 2010.

\bibitem[Noroozi \& Favaro(2016)Noroozi and Favaro]{noroozi2016unsupervised}
Noroozi, M. and Favaro, P.
\newblock Unsupervised learning of visual representations by solving jigsaw
  puzzles.
\newblock In \emph{European Conference on Computer Vision}, pp.\  69--84.
  Springer, 2016.

\bibitem[Nowozin et~al.(2016)Nowozin, Cseke, and Tomioka]{fgan}
Nowozin, S., Cseke, B., and Tomioka, R.
\newblock f-gan: Training generative neural samplers using variational
  divergence minimization.
\newblock In \emph{Advances in Neural Information Processing Systems}, pp.\
  271--279, 2016.

\bibitem[Palmer et~al.(2015)Palmer, Marre, Berry, and
  Bialek]{palmer2015predictive}
Palmer, S.~E., Marre, O., Berry, M.~J., and Bialek, W.
\newblock Predictive information in a sensory population.
\newblock \emph{Proceedings of the National Academy of Sciences}, 112\penalty0
  (22):\penalty0 6908--6913, 2015.

\bibitem[Paninski(2003)]{paninski2003estimation}
Paninski, L.
\newblock Estimation of entropy and mutual information.
\newblock \emph{Neural computation}, 15\penalty0 (6):\penalty0 1191--1253,
  2003.

\bibitem[Peng et~al.(2018)Peng, Kanazawa, Toyer, Abbeel, and
  Levine]{peng2018variational}
Peng, X.~B., Kanazawa, A., Toyer, S., Abbeel, P., and Levine, S.
\newblock Variational discriminator bottleneck: Improving imitation learning,
  inverse rl, and gans by constraining information flow.
\newblock \emph{arXiv preprint arXiv:1810.00821}, 2018.

\bibitem[Pfau \& Burgess(2018)Pfau and Burgess]{pfau2018minimally}
Pfau, D. and Burgess, C.~P.
\newblock Minimally redundant laplacian eigenmaps.
\newblock 2018.

\bibitem[Poole et~al.(2016)Poole, Alemi, Sohl-Dickstein, and
  Angelova]{poole2016}
Poole, B., Alemi, A.~A., Sohl-Dickstein, J., and Angelova, A.
\newblock Improved generator objectives for gans.
\newblock \emph{arXiv preprint arXiv:1612.02780}, 2016.

\bibitem[Rainforth et~al.(2018)Rainforth, Cornish, Yang, and
  Warrington]{rainforth2018nesting}
Rainforth, T., Cornish, R., Yang, H., and Warrington, A.
\newblock On nesting monte carlo estimators.
\newblock In \emph{International Conference on Machine Learning}, pp.\
  4264--4273, 2018.

\bibitem[Reshef et~al.(2011)Reshef, Reshef, Finucane, Grossman, McVean,
  Turnbaugh, Lander, Mitzenmacher, and Sabeti]{reshef2011detecting}
Reshef, D.~N., Reshef, Y.~A., Finucane, H.~K., Grossman, S.~R., McVean, G.,
  Turnbaugh, P.~J., Lander, E.~S., Mitzenmacher, M., and Sabeti, P.~C.
\newblock Detecting novel associations in large data sets.
\newblock \emph{science}, 334\penalty0 (6062):\penalty0 1518--1524, 2011.

\bibitem[Rezende et~al.(2014)Rezende, Mohamed, and
  Wierstra]{rezende2014stochastic}
Rezende, D.~J., Mohamed, S., and Wierstra, D.
\newblock Stochastic backpropagation and approximate inference in deep
  generative models.
\newblock In \emph{International Conference on Machine Learning}, pp.\
  1278--1286, 2014.

\bibitem[Ryan et~al.(2016)Ryan, Drovandi, McGree, and Pettitt]{ryan2016review}
Ryan, E.~G., Drovandi, C.~C., McGree, J.~M., and Pettitt, A.~N.
\newblock A review of modern computational algorithms for bayesian optimal
  design.
\newblock \emph{International Statistical Review}, 84\penalty0 (1):\penalty0
  128--154, 2016.

\bibitem[Saxe et~al.(2018)Saxe, Bansal, Dapello, Advani, Kolchinsky, Tracey,
  and Cox]{michael2018on}
Saxe, A.~M., Bansal, Y., Dapello, J., Advani, M., Kolchinsky, A., Tracey,
  B.~D., and Cox, D.~D.
\newblock On the information bottleneck theory of deep learning.
\newblock In \emph{International Conference on Learning Representations}, 2018.

\bibitem[Tishby \& Zaslavsky(2015)Tishby and Zaslavsky]{tishby2015deep}
Tishby, N. and Zaslavsky, N.
\newblock Deep learning and the information bottleneck principle.
\newblock In \emph{Information Theory Workshop (ITW), 2015 IEEE}, pp.\  1--5.
  IEEE, 2015.

\bibitem[Tishby et~al.(2000)Tishby, Pereira, and Bialek]{tishby2000information}
Tishby, N., Pereira, F.~C., and Bialek, W.
\newblock The information bottleneck method.
\newblock \emph{arXiv preprint physics/0004057}, 2000.

\bibitem[Tomczak \& Welling(2018)Tomczak and Welling]{pmlr-v84-tomczak18a}
Tomczak, J. and Welling, M.
\newblock Vae with a vampprior.
\newblock In Storkey, A. and Perez-Cruz, F. (eds.), \emph{Proceedings of the
  Twenty-First International Conference on Artificial Intelligence and
  Statistics}, volume~84 of \emph{Proceedings of Machine Learning Research},
  pp.\  1214--1223, Playa Blanca, Lanzarote, Canary Islands, 09--11 Apr 2018.
  PMLR.

\bibitem[van~den Oord et~al.(2016)van~den Oord, Kalchbrenner, Espeholt,
  Vinyals, Graves, et~al.]{van2016conditional}
van~den Oord, A., Kalchbrenner, N., Espeholt, L., Vinyals, O., Graves, A.,
  et~al.
\newblock Conditional image generation with pixelcnn decoders.
\newblock In \emph{Advances in Neural Information Processing Systems}, pp.\
  4790--4798, 2016.

\bibitem[van~den Oord et~al.(2018)van~den Oord, Li, and
  Vinyals]{oord2018representation}
van~den Oord, A., Li, Y., and Vinyals, O.
\newblock Representation learning with contrastive predictive coding.
\newblock \emph{arXiv preprint arXiv:1807.03748}, 2018.

\end{thebibliography}
\bibliographystyle{icml2019}
\vfill
\onecolumn
\begin{appendix}
\section{Summary of mutual information lower bounds}
In Table~\ref{table:comp}, we summarize the characteristics of lower bounds on MI. The parameters and objectives used for each of these bounds is presented in Table~\ref{table:eqns}.

\begin{table*}[h!]
\begin{center}
\begin{tabular}{ll|l|l|l|l|l}
 \multicolumn{2}{c|}{Lower Bound}& $L$ & $\nabla L$ & $\perp$ BS & Var. & Norm.\\
\hline
   $\ba$&~\citet{barber2003algorithm} & \xmark & \cmark & \cmark & \cmark & \xmark\\
   $\dv$&~\citet{donsker1983asymptotic}  & \xmark  &   \xmark       &      --  &      -- &--    \\
   $\inwj$&~\citet{nguyen2010estimating}   &    \cmark      &    \cmark      & \cmark                 &  \xmark        & \cmark\\
   $\mine$&~\citet{belghazi2018mutual}  & \xmark  &   \cmark       &        \cmark    &      \xmark &\cmark    \\
   $\cpc$&~\citet{oord2018representation}      &    \cmark      &   \cmark       &      \xmark     &        \cmark  & \cmark\\
   $\ijs$& Appendix~\ref{app:ijs}& \cmark  &   \cmark       &        \cmark    &      \xmark &\cmark    \\
   $I_\alpha$&Eq.~\ref{eq:interp}& \cmark  &   \cmark       &        \xmark    &      \cmark &\cmark    \\
\end{tabular}
\vspace{2mm}
\caption{Characterization of mutual information lower bounds. Estimators can have a tractable (\cmark) or intractable (\xmark) objective ($L$), tractable (\cmark) or intractable (\xmark) gradients ($\nabla L$), be dependent (\xmark) or independent (\cmark) of batch size ($\perp$ BS), have high (\xmark) or low (\cmark)  variance (Var.), and requires a normalized (\xmark) vs unnormalized (\cmark) critic (Norm.).}

\label{table:comp}
\end{center}
\end{table*}

\begin{table*}[h!]
\begin{center}
\begin{tabular}{l|l|l}
Lower Bound & Parameters & Objective\\
\hline
   $\ba$& $q(x|y)$ tractable decoder &  $\E_{p(x, y)} \left[ \log q(x|y)  - \log p(x) \right]$ \\
   $\dv$& $f(x,y)$ critic& $\E_{p(x,y)}\left[\log f(x,y)\right] - \log\left(\E_{p(x)p(y)}\left[f(x,y)\right]\right)$\\
   $\inwj$& $f(x,y)$ &   $\E_{p(x,y)}\left[\log f(x,y)\right] - \frac{1}{e}\E_{p(x)p(y)}\left[f(x,y)\right]$ \\
   $\mine$& $f(x,y)$, $\text{EMA}(\log f)$ & $\dv$ for evaluation, $\igb(f, \text{EMA}(\log f))$ for gradient \\
   $\cpc$& $f(x,y)$  &  $\mathbb{E}_{p^K(x,y)}\left[\frac{1}{K}\sum_{i=1}^K \log \frac{f(y_i, x_i)}{\frac{1}{K}\sum_{j=1}^K f(y_i,x_j)}\right] $ \\
   $\ijs$& $f(x,y)$& $\inwj$ for evaluation, $f$-GAN JS for gradient\\
   $\igb$& $f(x,y)$, $a(y) > 0$& $\E_{p(x, y)} \left[ \log f(x,y) \right] - \E_{p(y)} \left[ \frac{\E_{p(x)} \left[ f(x , y) \right]}{a(y)} + \log(a(y)) - 1  \right]$ \\
   $\tcpc$& $e(y|x)$ tractable encocder & $\cpc$ with $f(x,y) = e(y|x)$\\
   $I_\alpha$& $f(x,y)$, $\alpha$, $q(y)$&
   $1+\E_{p(x_{1:K}, y)}\left[\log \frac{e^{f(x_1,y)}}{\alpha m(y; x_{1:K}) + (1-\alpha)q(y)}\right]$\\
\multicolumn{2}{c}{}&\qquad$-\E_{p(x_{1:K})p(y)}\left[ \frac{e^{f(x_1,y)}}{\alpha m(y; x_{1:K}) + (1-\alpha)q(y)}\right] 
$
\end{tabular}
\vspace{2mm}
\caption{Parameters and objectives for mutual information estimators.}
\label{table:eqns}
\end{center}
\end{table*}

\section{Experimental details}
\label{app:toy}
{\bf Dataset}. For each dimension, we sampled $(x_i, y_i)$ from a correlated Gaussian with mean 0 and correlation of $\rho$. We used a dimensionality of 20, i.e. $x \in \mathbb{R}^{20}, y \in \mathbb{R}^{20}$. Given the correlation coefficient $\rho$, and dimensionality $d=20$, we can compute the true mutual information: $I(x, y) = -\frac{d}{2} \log(1-\rho^2)$. For Fig.~\ref{fig:estimation}, we increase $\rho$ over time to show how the estimator behavior depends on the true mutual information. 

{\bf Architectures.} We experimented with two forms of architecture: separable and joint. Separable architectures independently mapped $x$ and $y$ to an embedding space and then took the inner product, i.e. $f(x,y) = h(x)^Tg(y)$ as in~\citep{oord2018representation}. Joint critics concatenate each $x,y$ pair before feeding it into the network, i.e. $f(x,y) = h([x, y])$ as in~\citep{belghazi2018mutual}. In practice, separable critics are much more efficient as we only have to perform $2N$ forward passes through neural networks for a batch size of $N$ vs. $N^2$ for joint critics. All networks were fully-connected networks with ReLU activations.

\section{Additional experiments}
\subsection{Exhaustive hyperparameter sweep.}
To better evaluate the tradeoffs between different bounds, we performed more extensive experiments on the toy problems in Fig.~\ref{fig:estimation}.
For each bound, we optimized over learning rate, architecture (separable vs. joint critic, number of hidden layers (1-3), hidden units per layer (256, 512, 1024, 2048), nonlinearity (ReLU or Tanh), and batch size (64, 128, 256, 512). In Table~\ref{tab:sweep}, we present the estimate of the best-performing hyperparameters for each technique.  For both the Gaussian and Cubic problem, $I_\alpha$ outperforms all approaches at all levels of mutual information between $X$ and $Y$. While the absolute estimates are improved after this hyperparameter sweep, the ordering of the approaches is qualitatively the same as in Fig.~\ref{fig:estimation}. We also experimented with the bounds that leverage known conditional distribution, and found that Eq.~\ref{eq:tnwj} that leverages a known $p(y|x)$ is highly accurate as it only has to learn the marginal $q(y)$.

\subsection{Effective bias-variance tradeoffs with $\interp$}
\begin{figure}[h!]
    \centering
    \includegraphics[width=0.7\columnwidth]{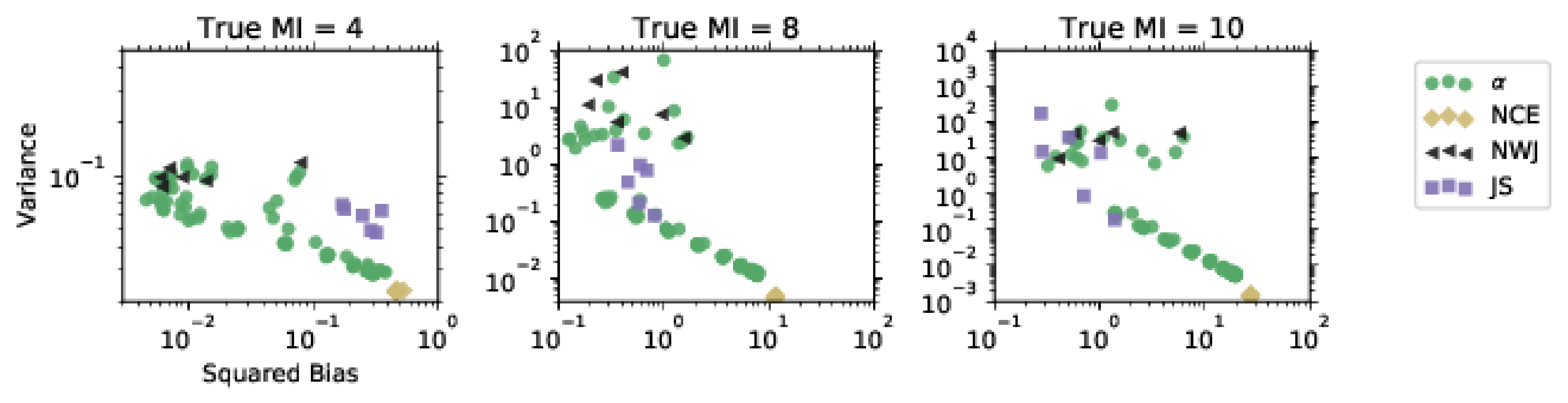}
    \caption{$I_\alpha$ effectively interpolates between $\cpc$ and $\inwj$, trading off bias for variance.}
    \label{fig:suppbiasvar}
\end{figure}

To better understand the effectiveness of $\interp$ at trading off bias for variance, we plotted bias vs. variance for 3 levels of mutual information on the toy 20-dimensional Gaussian problem across a range of architecture settings. In Fig.~\ref{fig:suppbiasvar}, we see that $\interp$ is able to effectively interpolate between the high-bias low-variance $\cpc$, and the low-bias high-variance $\inwj$. We also find that $\ijs$ is competitive at high rates, but exhibits higher bias and variance than $\interp$ at lower rates.
\begin{table}[]
    \centering
    \begin{tabular}{r|rrrrr}
    \multicolumn{1}{c}{}&\multicolumn{5}{c}{Mutual Information}\\
    \multicolumn{1}{c}{}& 2.0 & 4.0 & 6.0 & 8.0 & 10.0\\
    \midrule
    \multicolumn{1}{l}{\em{Gaussian, unstructured}}\\
 $\interp$  & 1.9 & 3.8 & 5.7 & 7.4 &  8.9 \\
 $\cpc$ & 1.9 & 3.6 & 4.9 & 5.7 &  6.0 \\
 $I_\text{JS}$  & 1.2 & 3.0 & 4.8 & 6.5 &  8.1 \\
 $\inwj$     & 1.6 & 3.5 & 5.2 & 6.7 &  8.0 \\
    \midrule
    \multicolumn{1}{l}{\em{Cubic, unstructured}}\\
$I_\alpha$& 1.7& 3.6& 5.4& 6.9& 8.2\\
$\cpc$& 1.7& 3.2& 4.1& 4.6& 4.8\\
$I_\text{JS}$& 1.0& 2.8& 4.5& 6.1& 7.6\\
$\inwj$& 1.5& 3.2& 4.7& 5.9& 6.9\\
\midrule
    \multicolumn{1}{l}{\em{Gaussian, known $p(y|x)$}}\\
    $\cpc$ (Eq.~\ref{eq:tcpc})&  1.9& 3.3& 4.2& 4.6& 4.8\\
    $\inwj$ (Eq.~\ref{eq:tnwj})&  2.0& 4.0& 6.0& 8.0& 10.0\\
    \end{tabular}
    \caption{Hyperparameter-optimizes results on the toy Gaussian and Cubic problem of Fig.~\ref{fig:estimation}. }
    \label{tab:sweep}
\end{table}

\begin{figure}[t!]
    \centering
    \includegraphics[width=0.7\columnwidth]{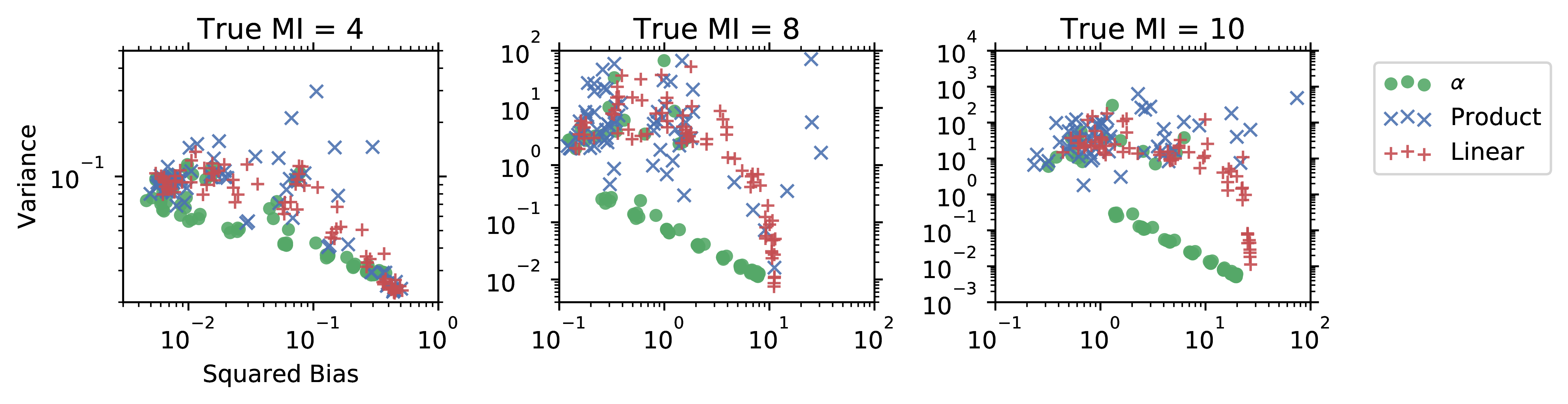}
    \caption{Comparing $I_\alpha$ to other interpolations schemes.}
    \label{fig:interpcomp}
\end{figure}

In addition to $\interp$, we compared to two alternative interpolation procedures, neither of which showed the improvements of $I_\alpha$:
\begin{enumerate}
    \item $\interp$ interpolation: multisample bound that uses a critic with linear interpolation between the batch mixture $m(y; x_{1:K}$ and the learned marginal $q(y)$ in the denominator (Eqn.~\ref{eq:interp}).
    \item Linear interpolation: $\alpha \cpc + (1-\alpha) \nwj$
    \item Product interpolation: same as $I_\alpha$, but uses the product $m(y;x_{1:K})^\alpha q(y)^{(1-\alpha)}$ in the denominator. 
\end{enumerate}
We compared these approaches in the same setting as Fig.~\ref{fig:suppbiasvar}, evaluating the bias and variance for various hyperparameter settings at three different levels of mutual information. In Fig.~\ref{fig:interpcomp}, we can see that neither the product or linear interpolation approaches reduce the bias or variance as well as $I_\alpha$.

\section{$\ijs$ derivation}
\label{app:ijs}
Given the high-variance of $\inwj$,  optimizing the critic with this objective can be challenging. Instead, we can optimize the critic using the lower bound on Jensen-Shannon (JS) divergence as in GANs and \citet{hjelm2018learning}, and use the density ratio estimate from the JS critic to construct a critic for the KL lower bound. 

The optimal critic for $\inwj$/$f$-GAN KL that saturates the lower bound on $KL(p\|q)$ is given by~\citep{fgan}:
$$T^*(x) = 1 + \log \frac{p(x)}{q(x)}.$$
If we use the $f$-GAN formulation for parameterizing the critic with a softplus activation, then we can read out the density ratio from the real-valued logits $V(x)$:
 $$\frac{p(x)}{q(x)} \approx \text{exp}\left({V(x)}\right)$$
 In~\citet{poole2016, mescheder2017adversarial}, they plug in this estimate of the density ratio into a Monte-Carlo approximation of the $f$-divergence. However, this is no longer a bound on the $f$-divergence, it is just an approximation. Instead, we can construct a critic for the KL divergence, $T_{KL}(x) = 1 + V(x)$, and use that to get a lower bound using the $\inwj$ objective:
\begin{align}
KL(p \| q) &\ge \E_{x \sim p}\left[T_{KL}(x)\right] - \mathbb{E}_{x \sim q}\left[\text{exp}(T_{KL}(x)-1)\right] \\
&= 
1 + \E_{x \sim p}\left[V(x)\right] - \mathbb{E}_{x \sim q}\left[\text{exp}(V(x))\right] 
\end{align}
Note that if the log density ratio estimate $V(x)$ is exact, i.e. $V(x) = \log\frac{p(x)}{q(x)}$, then the last term,  $\E_{x \sim q}\left[\text{exp}(V(x))\right]$ will be one, and the first term is exactly $KL(p \| q)$.

For the special case of mutual information estimation, $p$ is the joint $p(x,y)$ and $q$ is the product of marginals $p(x)p(y)$, yielding:
\begin{align}
I(X; Y) &\ge 1 + \E_{p(x,y)}\left[V(x, y)\right] - \E_{p(x)p(y)}\left[\text{exp}(V(x, y))\right] \triangleq \ijs.
\end{align}
\section{Alternative derivation of $\tcpc$}
In the main text, we derive $\cpc$ (and $\tcpc$) from a multi-sample variational lower bound. Here we present a simpler and more direct derivation of $\tcpc$. Let $p(x)$ be the data distribution, and $p(x_{1:K})$ denote $K$ samples drawn iid from $p(x)$.  Let $p(y|x)$ be a stochastic encoder, and $p(y)$ be the intractable marginal $p(y) = \int dx\, p(x) p(y|x)$. First, we can write the mutual information as a sum over K terms each of whose expectation is the mutual information:
\begin{align}
I(X; y) &= \E_{x_{1:K}}\left[\frac{1}{K} \sum_{i=1}^K \kl(p(y|x_i) \| p(y))\right]=  \E_{x_{1:K}}\left[\frac{1}{K}\sum_{i=1}^K \int dy\, p(y|x_i) \log \frac{p(y|x_i)}{p(y)}\right]\\
\end{align}
Let $m(y; x_{1:K}) = \frac{1}{K} \sum_{i=1}^K  p(y|x_i)$ be the minibatch estimate of the intractable marginal $p(y)$.  We multiply and divide by $m$ and then simplify:
\begin{align}
I(X; y) &= \E_{x_{1:K}}\left[\frac{1}{K}\sum_{i=1}^K \int dy\, p(y|x_i) \log \frac{p(y|x_i) m(y; x_{1:K})}{ m(y; x_{1:K}) p(y)}\right]\\
 &=\E_{x_{1:K}}\left[\frac{1}{K}\sum_{i=1}^K \left[\int dy\, p(y|x_i) \log \frac{p(y|x_i)}{ m(y; x_{1:K})} + \int dy\, p(y|x_i) \log \frac{m(y; x_{1:K})}{p(y)}\right]\right]\\
 &=\E_{x_{1:K}}\left[\frac{1}{K}\sum_{i=1}^K \kl(p(y|x_i) \| m(y; x_{1:K}))  + \ \int dy\, \frac{1}{K} \sum_{i=1}^K p(y|x_i) \log \frac{m(y; x_{1:K})}{p(y)}\right]\\
 &=\E_{x_{1:K}}\left[\left(\frac{1}{K}\sum_{i=1}^K \kl(p(y|x_i) \| m(y; x_{1:K}))\right) + \kl (m(y; x_{1:K}) \| p(y))\right]\\
 &\ge\E_{x_{1:K}}\left[\frac{1}{K}\sum_{i=1}^K \kl(p(y|x_i) \| m(y; x_{1:K}))\right]
\end{align}

\end{appendix}
\end{document}